\documentclass{llncs}

\usepackage{color,xcolor}
\usepackage{graphicx}
\usepackage{caption}
\usepackage{subcaption}
\usepackage{multicol}
\usepackage{verbatim}
\usepackage{array,multirow}
\usepackage{epsfig}
\usepackage{rotating}
\usepackage{amsmath}

\usepackage{tikz}
\usetikzlibrary{arrows}
\usetikzlibrary{shapes}

\newcolumntype{C}[1]{>{\centering\let\newline\\\arraybackslash\hspace{0pt}}m{#1}}

\begin{document}

\title{Dense Bag-of-Temporal-SIFT-Words\\ for Time Series Classification}
\author{Adeline Bailly\inst{1,4} \and Simon Malinowski\inst{2} \and
Romain Tavenard\inst{1} \and \\ Laetitia Chapel\inst{3} \and Thomas Guyet\inst{4}
}
\institute{Universit{\'e} de Rennes 2, IRISA, LETG-Rennes COSTEL, Rennes, France 
\and Universit{\'e} de Rennes 1, IRISA, Rennes, France
\and Universit{\'e} de Bretagne-Sud, IRISA, Vannes, France
\and Agrocampus Ouest, IRISA, Rennes, France
}

\maketitle

\setcounter{footnote}{0}

\begin{abstract}
The SIFT framework has shown to be accurate in the image classification context. In~\cite{botsw15}, we designed a Bag-of-Words approach based on an adaptation of this framework to time series classification. It relies on two steps: SIFT-based features are first extracted and quantized into words; histograms of occurrences of each word are then fed into a classifier. In this paper, we investigate techniques to improve the performance of Bag-of-Temporal-SIFT-Words: dense extraction of keypoints and normalization of Bag-of-Words histograms. Extensive experiments show that our method significantly outperforms most state-of-the-art techniques for time series classification.
\keywords{time series classification, Bag-of-Words, SIFT, dense features, BoTSW, D-BoTSW}
\end{abstract}

\section{Introduction \label{sec:intro} }

Classification of time series has received an important amount of interest over the past years due to many real-life applications, such as medicine~\cite{wang2012bowbiomedical}, environmental modeling~\cite{dusseux13}, speech recognition~\cite{lecun95}. 
%
%
A wide range of algorithms have been proposed to solve this problem.
One simple classifier is the $k$-nearest-neighbor ($k$NN), which is usually combined with Euclidean Distance (ED) or Dynamic Time Warping (DTW) similarity measure. The combination of the $k$NN classifier with DTW
is one of the most popular method since it achieves high classification accuracy~\cite{ratanamahatana2004everything}. However, this method has a high computation cost which makes its use difficult for large-scale real-life applications.

Above-mentioned techniques compute similarity between time series based on point-to-point comparisons. 
Classification techniques based on higher level structures (\emph{e.g.} feature vectors) are most of the time faster, while being at least as accurate as DTW-based classifiers. Hence, various works have investigated the extraction of local and global features in time series. 
Among these works, the Bag-of-Words (BoW) approach (also called Bag-of-Features) 
consists in representing documents using a histogram of word occurrences.
It is a very common technique in text mining, information retrieval and content-based image retrieval because of its simplicity and performance. For these reasons, it has been adapted to time series data in some recent works~\cite{Bay15,baydogan2013bof,lin2012bop,senin2013saxsvm,wang2012bowbiomedical}. Different kinds of features based on simple statistics, computed at a local scale, are used to create the words.

In the context of image retrieval and classification, scale-invariant descriptors have proved their accuracy.
Particularly, the Scale-Invariant Feature Transform (SIFT) framework has led to widely used descriptors~\cite{lowe2004distinctive}.
These descriptors are scale and rotation invariant while being robust to noise.
In~\cite{botsw15}, we built on this framework to design a BoW approach for time series classification where words correspond to quantized versions of local features. Features 
are built using the SIFT framework for both detection and description of the keypoints. 
This approach can be seen as an adaptation of~\cite{sivic2003video}, which uses SIFT features associated with visual words, to time series. 
In this paper, we improve our previous work by applying enhancement techniques for BoW approaches, such as dense extraction and BoW normalization. 
To  validate this, we conduct extensive experiments on a wide range of datasets.

This paper is organized as follows. Section~\ref{sec:related} summarizes related work, Section~\ref{sec:proposed} describes the proposed Bag-of-Temporal-SIFT-Words (BoTSW) method and its improved version (dense extraction and BoW normalization, D-BoTSW), and Section~\ref{sec:xp} reports experimental results. Finally, Section~\ref{sec:ccl} concludes and discusses future work.

\section{Related work \label{sec:related} }

Our approach for time series classification builds on two well-known methods in computer vision: local features are extracted from time series using a SIFT-based approach and a global representation of time series is produced using Bag-of-Words.
This section first introduces state-of-the-art distance-based methods in time series classification and then presents previous works that make use of Bag-of-Words approaches for time series classification.

\subsection{Distance-based time series classification}

Data mining community has, for long, investigated the field of time series classification. Early works focus on the use of dedicated similarity measures to assess similarity between time series. In~\cite{ratanamahatana2004everything}, Ratanamahatana and Keogh compare Dynamic Time Warping to Euclidean Distance when used with a simple $k$NN classifier. While the former benefits from its robustness to temporal distortions to achieve high accuracy, ED is known to have much lower computational cost.
Cuturi~\cite{cuturi2011fast} shows that, although DTW is well-suited to retrieval tasks since it focuses on the best possible alignment between time series, it fails at precisely quantifying dissimilarity between non-matching sequences (which is backed by the fact that DTW-derived kernel is not positive definite).
Hence, he introduces the Global Alignment Kernel that takes into account all possible alignments in order to produce a reliable similarity measure to be used at the core of standard kernel methods such as Support Vector Machines (SVM).
Lines and Bagnall~\cite{lines14} propose an ensemble classifier based on elastic distance measures (including DTW), named Proportional Elastic Ensemble (PROP). 
Instead of building classification decision on similarities between time series, Ye and Keogh~\cite{ye2009time} use a decision tree in which the partitioning of time series is performed with respect to the presence (or absence) of discriminant sub-sequences (named shapelets) in the series. Though accurate, the method is very computational demanding as building the decision tree requires one to check for all candidate shapelets. Douzal and Amblard~\cite{douzal2010pr} define a dedicated similarity measure for time series which is then used in a classification tree.


\subsection{Bag-of-Words for time series classification}


Inspired by text mining, information retrieval and computer vision communities, recent works have investigated the use of Bag-of-Words for time series classification~\cite{Bay15,baydogan2013bof,lin2012bop,senin2013saxsvm,wang2012bowbiomedical}.
These works are based on two main operations: converting time series into Bag-of-Words, and building a classifier upon this BoW representation.
Usually, standard techniques such as random forests, SVM, neural networks or $k$NN are used for the classification step. 
Yet, many different ways of converting time series into Bag-of-Words have been introduced. Among them, Baydogan \emph{et al.} ~\cite{baydogan2013bof} propose a framework to classify time series 
denoted TSBF where local features such as mean, variance and extremum values are computed on sliding windows. These features are then quantized into words using a codebook learned by a class probability estimate distribution. 
In~\cite{wang2012bowbiomedical}, discrete wavelet coefficients are extracted on sliding windows and then quantized into words using $k$-means. 
In~\cite{lin2012bop,senin2013saxsvm}, words are constructed using the Symbolic Aggregate approXimation (SAX) representation~\cite{Lin03} of time series. SAX symbols are extracted from time series and histograms of $n$-grams of these symbols are computed to form a Bag-of-Patterns (BoP). In \cite{senin2013saxsvm}, Senin and Malinchik combine SAX with Vector Space Model to form the SAX-VSM method. In~\cite{Bay15}, Baydogan and Runger design a symbolic representation of multivariate time series (MTS), called SMTS, where MTS are transformed into a feature matrix, whose rows are feature vectors containing a time index, the values and the gradient of time series at this time index (on all dimensions). Random samples of this matrix are given to decision trees whose leaves are seen as words. A histogram of words is output when the different trees are learned.

Local feature extraction has been investigated for long in the computer vision community.
One of the most powerful local feature for image is SIFT~\cite{lowe2004distinctive}.
It consists in detecting keypoints as extremum values of the the Difference-of-Gaussians (DoG) function and describing their neighborhoods using histograms of gradients.
Xie and Beigi~\cite{Xie09} use similar keypoint detection for time series. 
Keypoints are then described by scale-invariant features that characterize the shapes surrounding the extremum.
In~\cite{Can12}, extraction and description of time series keypoints in a SIFT-like framework is used to reduce the complexity of DTW: features are used to match anchor points from two different time series and prune the search space when searching for the optimal path for DTW.

In this paper, we build upon BoW of SIFT-based descriptors.
We propose an adaptation of SIFT to mono-dimensional signals that preserves their robustness to noise 
and their scale invariance.
We then use BoW to gather information from many local features into a single global one.
 
\section{Bag-of-Temporal-SIFT-Words (BoTSW) method
\label{sec:proposed} }

\begin{figure}[t]
\centering
\begin{multicols}{2}
\begin{subfigure}[b]{0.48\textwidth}
	\includegraphics[height=18mm,width=\textwidth]{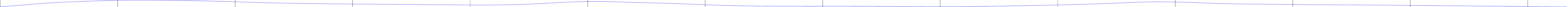}
	\includegraphics[height=9mm,width=\textwidth]{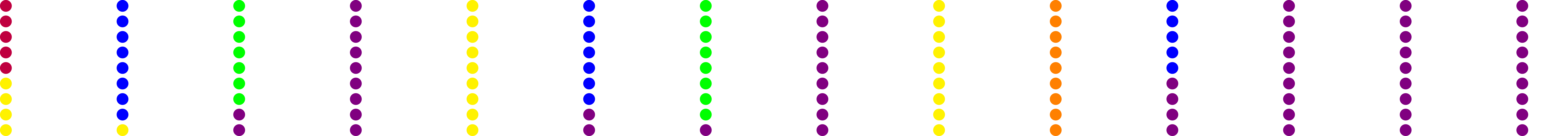}
    \caption{Dense extraction $(\tau_\text{step}=15, 9$ scales)}
\end{subfigure}
\par
\begin{subfigure}[b]{0.48\textwidth}
	\centering
	\includegraphics[height=22mm,width=0.32\textwidth]{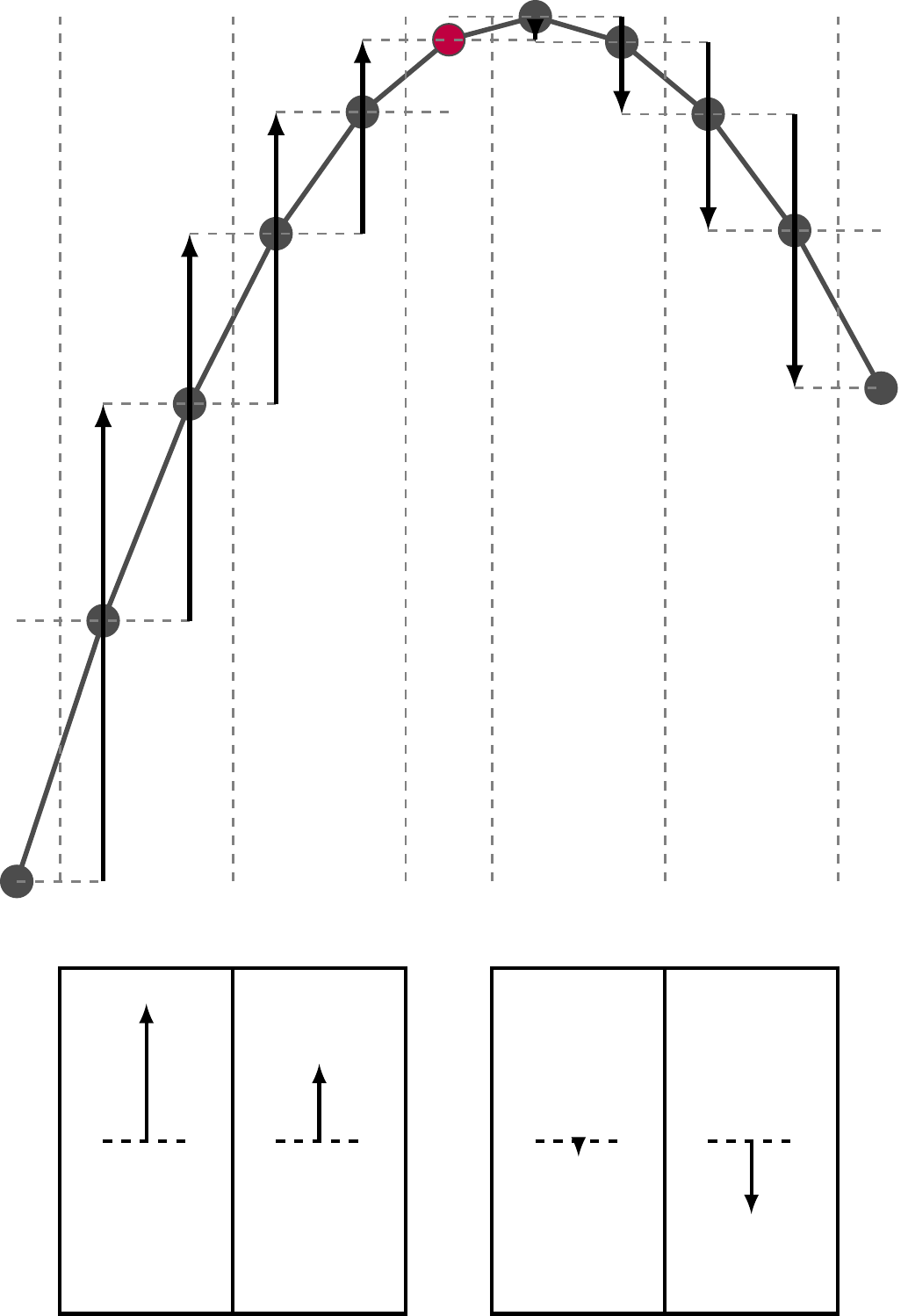}
	\includegraphics[height=22mm,width=0.32\textwidth]{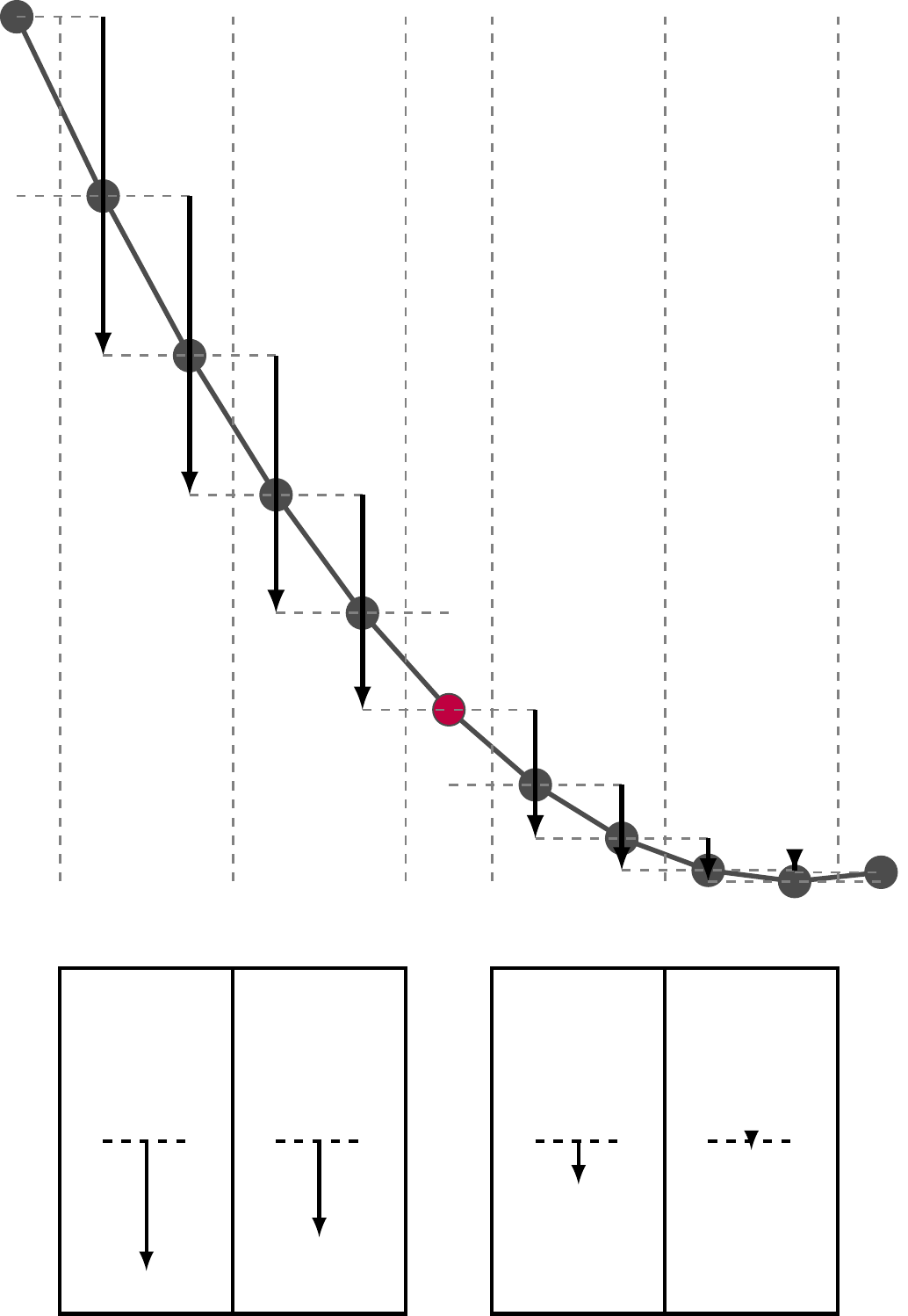}
	\includegraphics[height=22mm,width=0.32\textwidth]{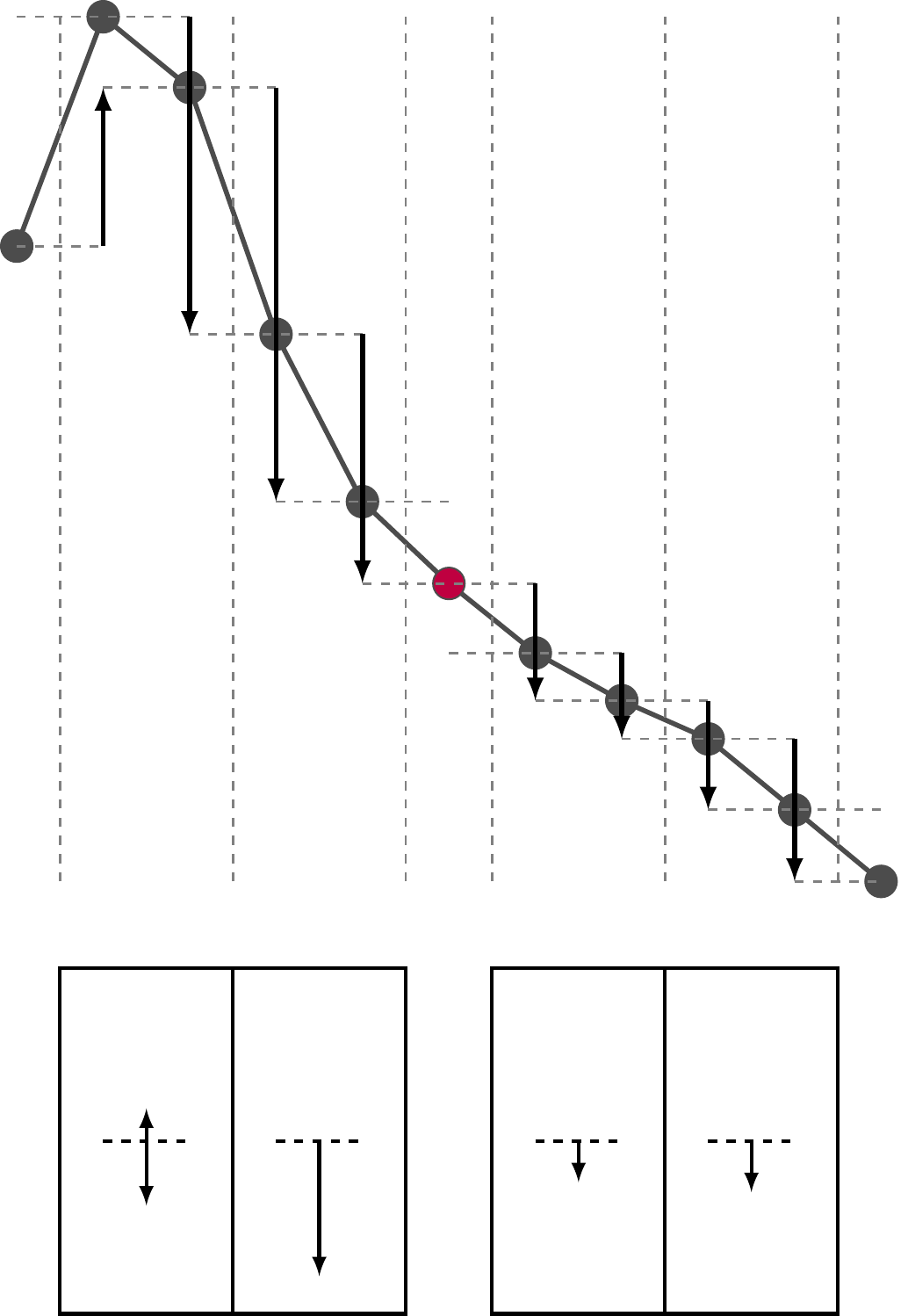}
    \caption{Keypoint description $(n_b=4, a=2)$}
\end{subfigure}
\par
\begin{subfigure}[b]{0.48\textwidth}
	\centering
	\includegraphics[height=9mm,width=0.45\textwidth]{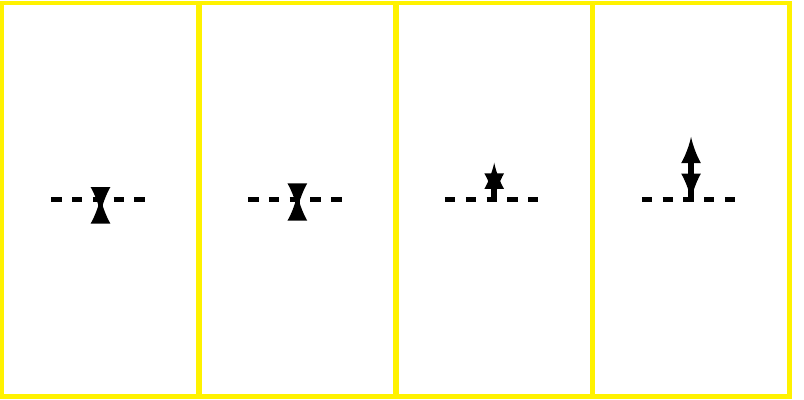}
	\includegraphics[height=9mm,width=0.45\textwidth]{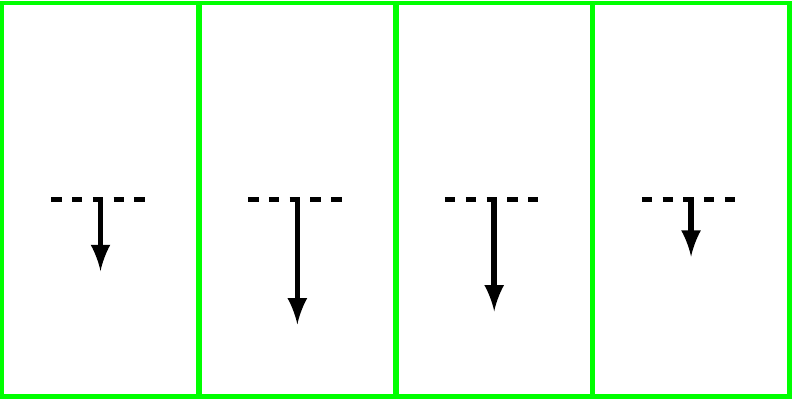}
	\includegraphics[height=9mm,width=0.45\textwidth]{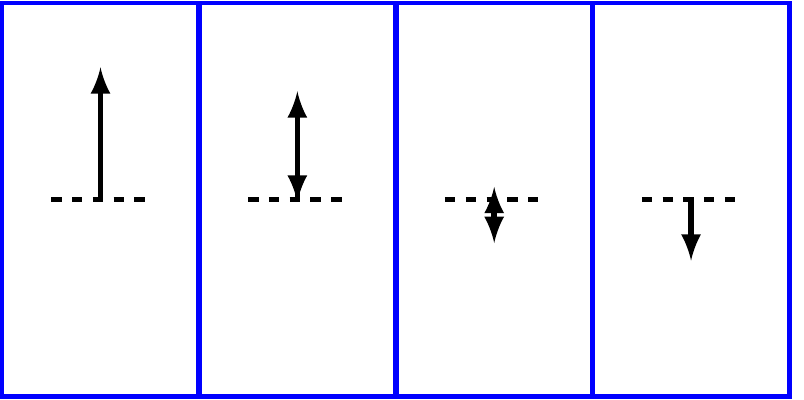}
	\includegraphics[height=9mm,width=0.45\textwidth]{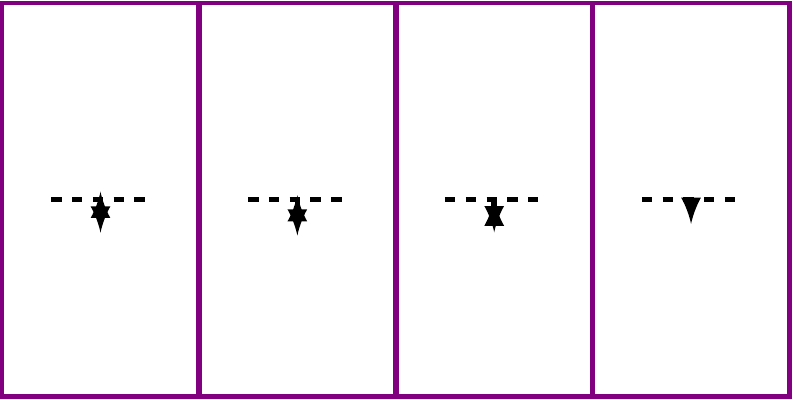}
	\includegraphics[height=9mm,width=0.45\textwidth]{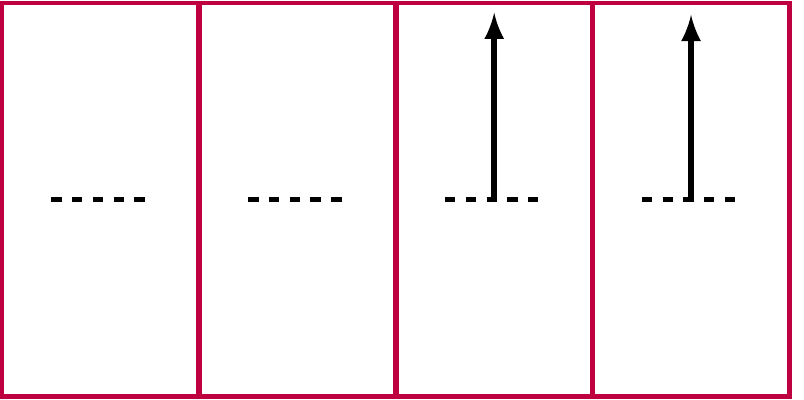}
	\includegraphics[height=9mm,width=0.45\textwidth]{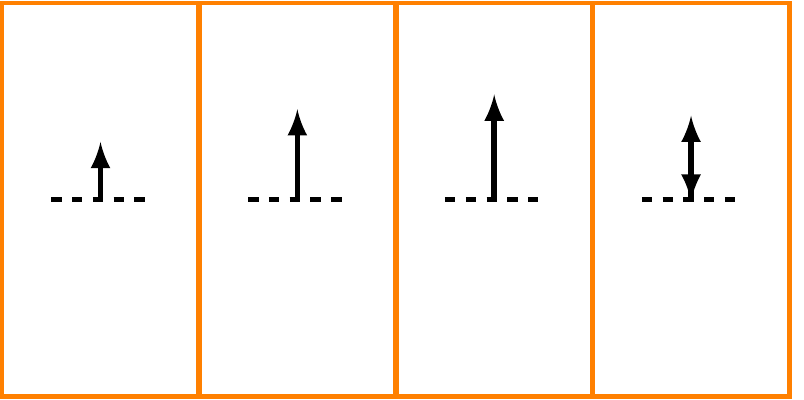}
    \caption{$k$-means generated codebook $(k=6)$}
\end{subfigure}
\par
\begin{subfigure}[b]{0.48\textwidth}
	\centering
	\includegraphics[height=22mm,width=\textwidth]{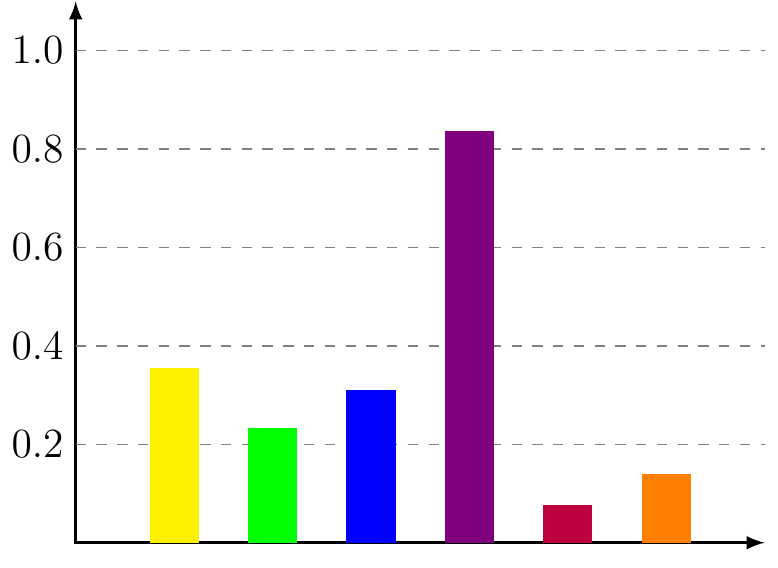}
    \caption{Resulting $k$-dimensional histogram}
\end{subfigure}
\end{multicols}
\caption{
Approach overview: (a) A time series and its dense-extracted keypoints. (b) Keypoint description is based on the time series filtered at the scale at which the keypoint is extracted. Descriptors are quantized into words. 
(c) Codewords obtained \emph{via} $k$-means, 
the color is associated with the dots under each keypoint in (a). (d) Histograms of word occurrences are given to a classifier (linear SVM) that learns boundaries between 
classes. Best viewed in color.\label{fig:overview}}
\end{figure}

The proposed method is 
based on three main steps: (i) extraction of keypoints in time series, (ii) description of these keypoints by gradient magnitude at a specific scale and (iii) representation of time series by a BoW, where words correspond to quantized version of the description of keypoints.
These steps are depicted in \figurename~\ref{fig:overview} and detailed below.

\subsection{Keypoints extraction in time series}

The first step of our method consists in extracting keypoints in time series. Two approaches are described here: the first one is based on scale-space extrema detection (as in~\cite{botsw15}) and the second one proposes a dense extraction scheme. 

\subsubsection{Scale-space extrema detection.}
Following the SIFT framework, keypoints in time series are detected as local extrema in terms of both scale and (temporal) location. These scale-space extrema are identified using a DoG function, and form a list of scale-invariant keypoints.
Let $L(t,\sigma)$ be the convolution ($\ast$) of a Gaussian function $G(t,\sigma)$ of width $\sigma$ with a time series $S(t)$:
\begin{equation}
L(t,\sigma) = G(t,\sigma) \ast S(t)
\end{equation}
where $G(t,\sigma)$ is defined as
\begin{equation}
G(t,\sigma) = \frac{1}{\sqrt{2\pi}~\sigma}~e^{- t^2 / 2\sigma^2}.
\end{equation}
Lowe~\cite{lowe1999objectrecognition} proposes the Difference-of-Gaussians (DoG) function to detect scale-space extrema in images. Adapted to time series, a DoG function is obtained by subtracting two time series filtered at consecutive scales:
\begin{equation}
D(t,\sigma) = L(t, k_\text{sc}\sigma) - L(t,\sigma),
\end{equation}
where $k_\text{sc}$ is a parameter of the method that controls the scale ratio between two  consecutive scales.

Keypoints are then detected at time index $t$ in scale $j$ if they correspond to extrema of $D(t, k_\text{sc}^j\sigma_0)$ in both time and scale, where $\sigma_0$ is the width of the Gaussian corresponding to the reference scale.
At a given scale, each point has two neighbors: one at the previous and one at the following time instant. 
Points also have neighbors one scale up and one scale down at the previous, same and next time instants, leading to a total of eight neighbors.
If a point is higher (or lower) than all of its neighbors, it is considered as an extremum in the scale-space domain and hence a keypoint of $S$. 

\subsubsection{Dense extraction.}
Previous researches have shown that accurate classification could be achieved by using densely extracted local features~\cite{jurie05,wang09}. 
In this section, we present the adaptation of this setup to our BoTSW scheme.
Keypoints selected with dense extraction no longer correspond to extrema but are rather systematically extracted at all scales every $\tau_\text{step}$ time steps on Gaussian-filtered time series $L(\cdot{},k_\text{sc}^j\sigma_0)$. 

Unlike scale-space extrema detection, regular sampling guarantees a minimal amount of keypoints per time series. 
This is especially crucial for smooth time series from which very few keypoints are detected when using scale-space extrema detection. In addition, even if the densely extracted keypoints are not scale-space extrema, description of these keypoints (cf. Section~\ref{ssec:desc}) covers the description of scale-space extrema if $\tau_\text{step}$ is not too large.
This usually leads to more robust global descriptors.

%
%


A dense extraction scheme is represented in~\figurename~\ref{fig:overview}, where we consider a step of $\tau_\text{step} = 15$ for the sake of readability. 
In the following, when dense extraction is performed, we will refer to our method as D-BoTSW (for dense BoTSW).

\subsection{Description of the extracted keypoints}
\label{ssec:desc}

Next step in our process is the description of keypoints. 
A keypoint at time index $t$ and scale $j$ is described by gradient magnitudes of $L(\cdot{}, k_\text{sc}^j\sigma_0)$ around $t$. 
To do so, $n_b$ blocks of size $a$ are selected around the keypoint. Gradients are computed at each point of each block and weighted using a Gaussian window of standard deviation $\frac{a \times n_b}{2}$ so that points that are farther in time from the detected keypoint have lower influence. 
Then, each block is described by two values: the sum of positive gradients and the sum of negative gradients. 
Resulting feature vector is hence of dimension $2 \times n_b$.

\subsection{Bag-of-Temporal-SIFT-Words for time series classification}
\label{botsw_v_features}

The set of all training features is used to learn a codebook of $k$ words using $k$-means clustering. 
Words represent different local behaviors in time series.
Then, for a given time series, each feature vector is assigned the closest word in the codebook.
The number of occurrences of each word in a time series is computed. 
(D-)BoTSW representation of a time series is the $\ell_2$-normalized histogram (\emph{i.e.} frequency vector) of word occurrences. 

\subsubsection{Bag-of-Words normalization.}
Dense sampling on multiple Gaussian-filtered time series provides considerable information to process.
It also tends to generate words with little informative power, as stop words do in text mining applications.
In order to reduce the impact of those words, we compare two normalization schemes for BoW: Signed Square Root normalization (SSR) and Inverse Document Frequency normalization (IDF). These normalizations are commonly used in image retrieval and classification based on histograms~\cite{jegou12,jegou2010aggregating,perronin10,sivic2003video}.

J{\'e}gou \emph{et al.}~\cite{jegou2010aggregating} and Perronin \emph{et al.}~\cite{perronin10} show that reducing the influence of frequent codewords before $\ell_2$ normalization could be profitable. 
They apply a power $\alpha \in [0,1]$ on their global representation. SSR normalization corresponds to the case where $\alpha = 0.5$, which leads to near-optimal results~\cite{jegou2010aggregating,perronin10}.

IDF normalization also tends to lower the influence of frequent codewords. 
To do so, document frequency of words is computed as the number of training time series in which the word occurs.
BoW are then updated by diving each component by its associated document frequency.

SSR and IDF normalizations both reduce the influence of frequent codewords in the codebook, and are applied before  $\ell_2$ normalization. 
We show in the experimental part of this paper that using BoW normalization improves the accuracy of our method.

Normalized histograms are finally given to a classifier that learns how to discriminate classes from this D-BoTSW representation.

\section{Experiments and results}
\label{sec:xp}

In this section, we investigate the impact of both dense extraction of the keypoints and normalization of the Bag-of-Words on classification performance. We then compare our results to the ones obtained with standard time series classification techniques.

For the sake of reproducibility, C++ source code used for (D-)BoTSW in these experiments is made available for download\footnote{ \url{http://people.irisa.fr/Adeline.Bailly/code.html}}.
%
%
To provide illustrative timings for our methods, we ran it on a personal computer, for a given set of parameters, using dataset \emph{Cricket\_X}~\cite{ucr} that is made of 390 training time series and 390 test ones.
Each time series in the dataset is of length 300.
Extraction and description of dense keypoints takes around 1 second for all time series in the dataset.
Then, 35 seconds are necessary to learn a $k$-means and fit a linear SVM classifier using training data only.
Finally, classification of all D-BoTSW corresponding to test time series takes less than 1 second.


\subsection{Experimental setup}
\label{sec:xp_setup}

Experiments are conducted on the 86 currently available datasets from the UCR repository~\cite{ucr}, the largest online database for time series classification. It includes a wide variety of problems, such as sensor reading (\emph{ECG}), image outline (\emph{ArrowHead}), human motion (\emph{GunPoint}), as well as simulated problems (\emph{TwoPatterns}). All datasets are split into a training and a test set, whose size varies between less than 20 and more than 8000 time series. 
For a given dataset, all time series have the same length, ranging from 24 to more than 2500 points. 

Parameters $a$, $n_b$, $k$ and $C_{SVM}$ of (D-)BoTSW are learned, while we set $\sigma_0 = 1.6$ and $k_\text{sc} = 2^{1/3}$, as these values have shown to produce stable results~\cite{lowe2004distinctive}.
Parameters $a$, $n_b$, $k$ and $C_{SVM}$ vary inside the following sets: $\{4, 8\}$, $\{4, 8, 12, 16, 20\}$, $\left\{2^i, \forall i \in \{5..10\}\right\}$ and $\{1, 10, 100\}$ respectively.
Codebooks are obtained \emph{via} $k$-means quantization and a linear SVM is used to classify time series represented as (D-)BoTSW. 
%
For our approach, the best sets (in terms of accuracy) of $(a, n_b, k, C_{SVM})$ parameters are selected by performing cross-validation on the training set. 
Due to the heterogeneity of the datasets, leave-one-out cross-validation is performed on datasets where the training set contains less than 300 time series, and $10$-fold cross-validation is used otherwise. 
These best sets of parameters are then used to build the classifier on the training set and evaluate it on the test set.
For datasets with little training data, it is likely that several sets of parameters yield best performance during the cross-validation process.
For example, when using \emph{DiatomSizeReduction} dataset, BoTSW has 150 out of 180 parameter sets yielding best performance, while there are 42 such sets for D-BoTSW with SSR normalization.
In both cases, the number of \emph{best} parameter sets is too high to allow a fair parameter selection.
When this happens, we keep all parameter sets with best performance at training and perform a majority voting between their outputs at test time.

Parameters $a$ and $n_b$ both influence the descriptions of the keypoints; their optimal values vary between sets so that the description of keypoints can fit the shape of the data. 
If the data contains sharp peaks, the size of the neighborhood on which features are computed (equal to $a\times{}n_b$) should be small.
On the contrary, if it contains smooth peaks, descriptions should take more points into account.
Parameter $k$ of the $k$-means needs to be large enough to precisely represent the different features.
However, it needs to be small enough in order to avoid overfitting.
We consequently allow a large range of values for $k$. 


In the following, BoTSW denotes the approach where keypoints are selected as scale-space extrema and BoW histograms are $\ell_2$-normalized.
For all experiments with dense extraction, we set $\tau_\text{step} = 10$, and we extract keypoints at all scales. Using such a value for $\tau_\text{step}$ enables one to have a sufficient number of keypoints even for small time series, and guarantees that keypoint neighborhoods overlap so that all subparts of the time series are described.

\subsection{Experiments on dense extraction}
\label{sec:xp_dense}

\begin{figure}[!ht]
\centering
	\includegraphics[width=0.48\textwidth]{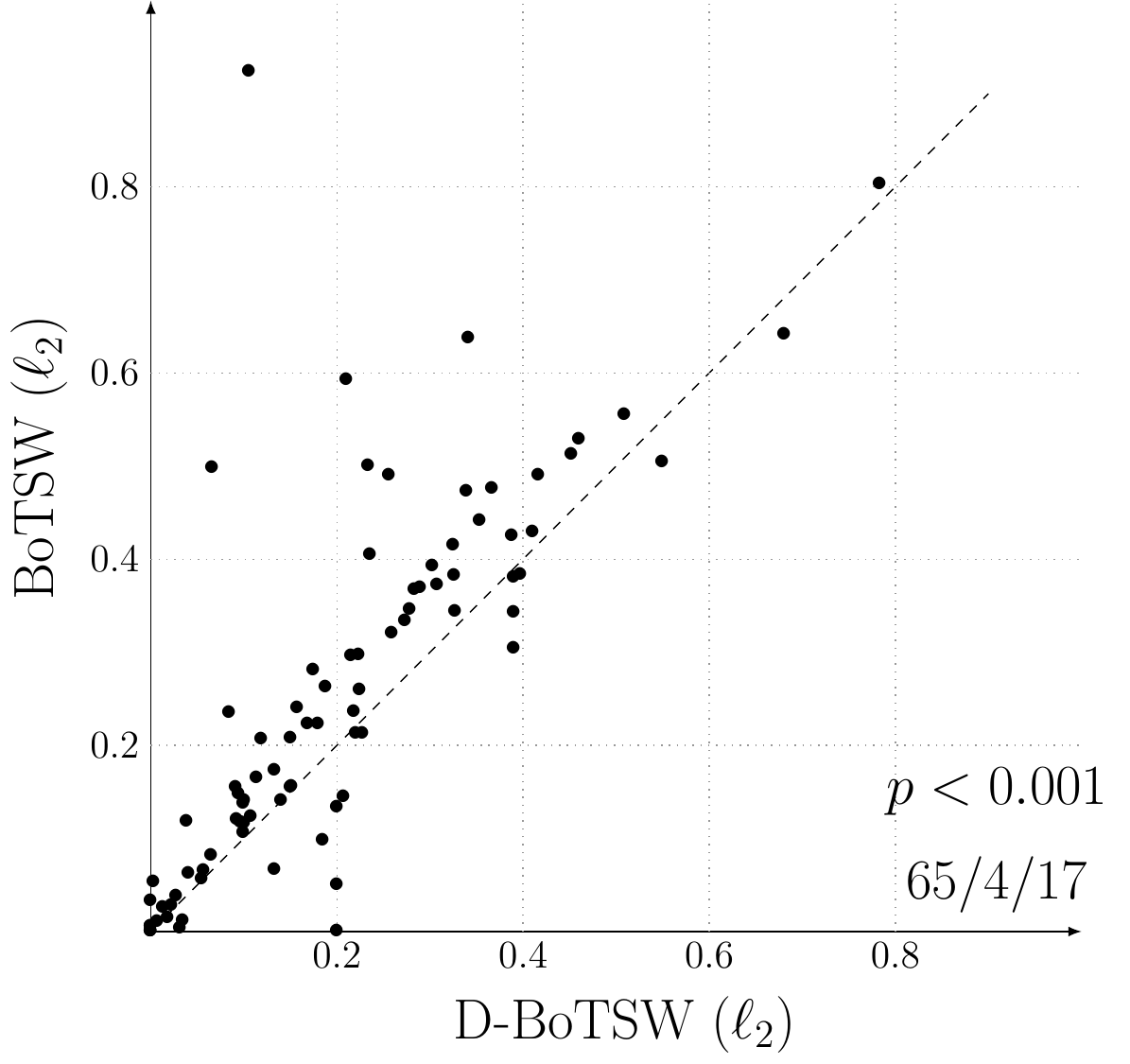}
\caption{\label{fig:errorrate_basic_dense}
Error rates of BoTSW compared to D-BoTSW.
}
\end{figure}

\figurename~\ref{fig:errorrate_basic_dense} shows a pairwise comparison of error rates between BoTSW and its dense counterpart D-BoTSW for all datasets in the UCR repository. 
A point on the diagonal means that obtained error rates are equals. A point above the diagonal illustrates a case where D-BoTSW has a smaller error rate than BoTSW.
Wilcoxon signed rank test's $p$-value and Win/Tie/Lose scores are given in the bottom-right corner of the figure.
Win/Tie/Lose scores indicate that D-BoTSW reaches better performance than BoTSW on 61 datasets, equivalent performance on 4 datasets and worse on 21 datasets.
Wilcoxon test shows that this difference is significant (in the following, we will use a significance level of 10\% for all statistical tests).

D-BoTSW improves classification on a large majority of the datasets.
However, most points are close to the diagonal, which means that the improvement is of little magnitude. 
In the following, we show how to further improve these results thanks to D-BoTSW normalization.

\subsection{Experiments on BoW normalization}
\label{sec:xp_norm}

\begin{figure}[!ht]
\centering
	\includegraphics[width=0.48\textwidth]{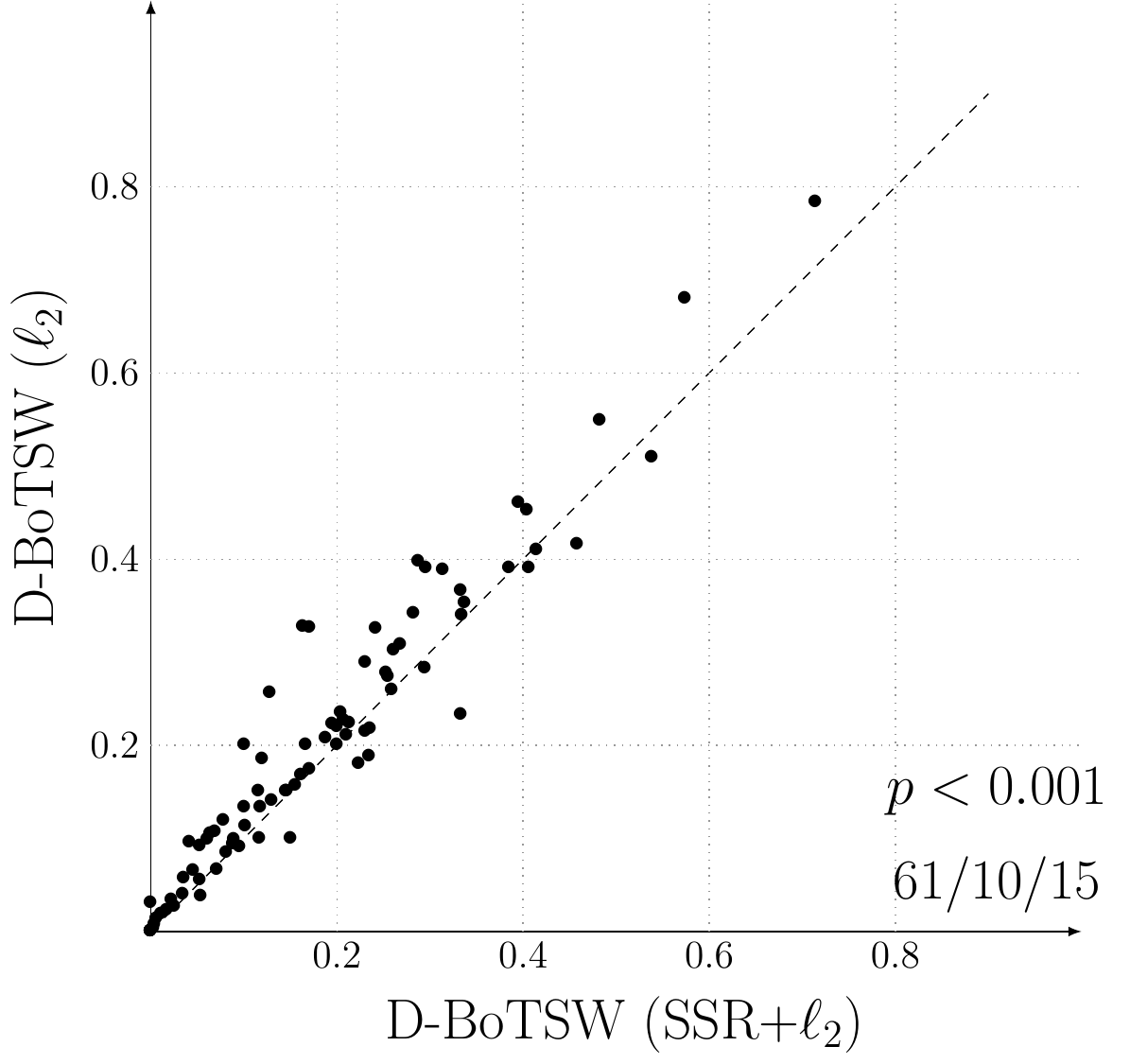}
	\includegraphics[width=0.48\textwidth]{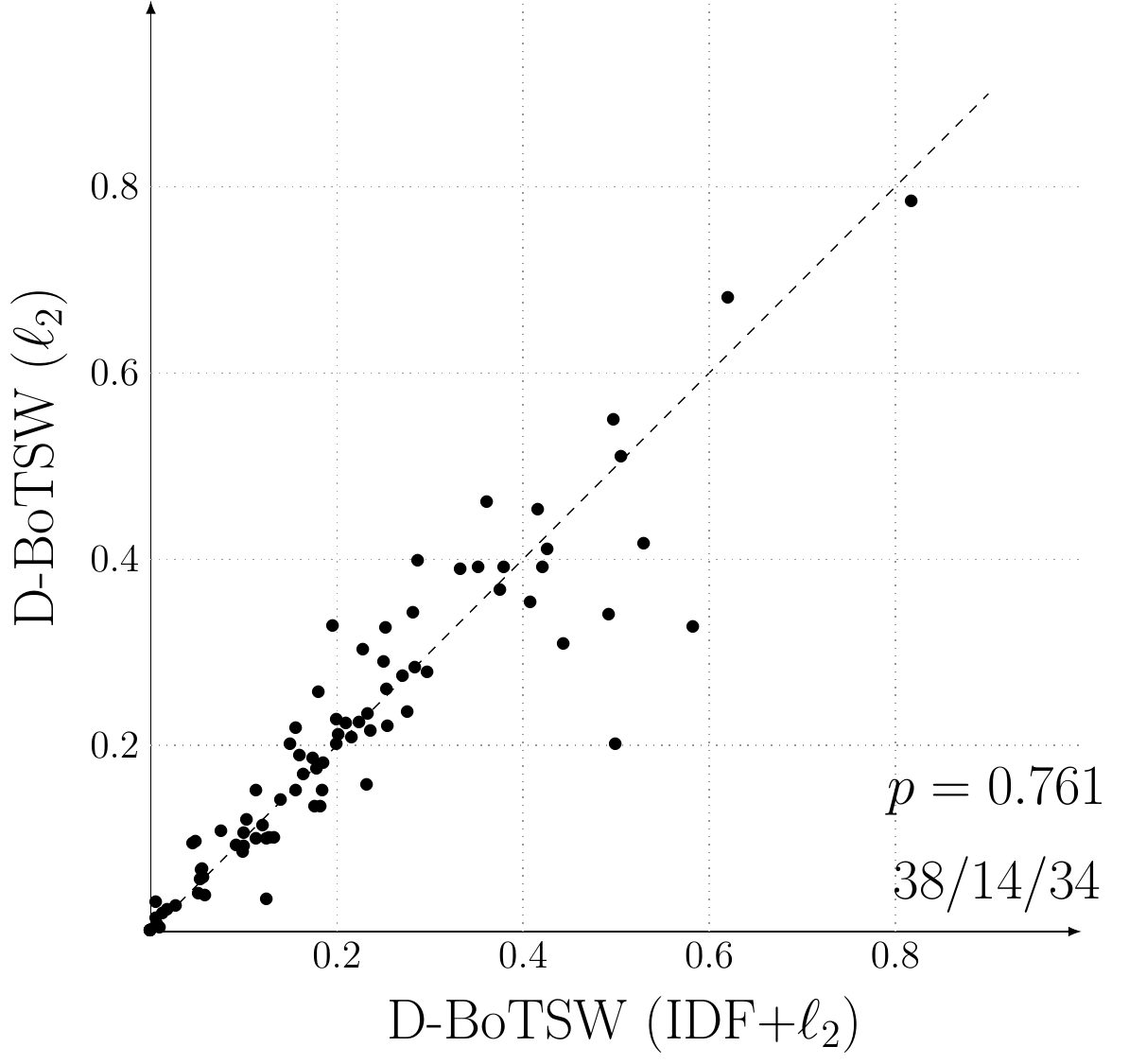}
\caption{\label{fig:errorrate_dense_norm}
Error rates of D-BoTSW with and without normalization.
}
\end{figure}

In image retrieval and classification, Bag-of-Words normalizations have been shown to improve classification rates with dense extracted keypoints. We investigate here the impact of SSR and IDF normalizations on D-BoTSW for time series classification.

As it can be seen in~\figurename~\ref{fig:errorrate_dense_norm}, both SSR and IDF normalizations improve classification performance (though the improvement of using IDF is not statistically significant). Lowering the influence of largely-represented codewords hence leads to more accurate classification with D-BoTSW.


%
%

IDF normalization only leads to a small improvement in classification accuracy: Win/Tie/Lose score against non-normalized D-BoSTW is 38/14/34. 
On the contrary, SSR normalization significantly improves the classification accuracy, with a Win/Tie/Lose score of 61/10/15 over non-normalized D-BoSTW.


\begin{figure}[t]
\centering
\begin{subfigure}[b]{0.8\textwidth}
	\centering
	\includegraphics[width=\linewidth]{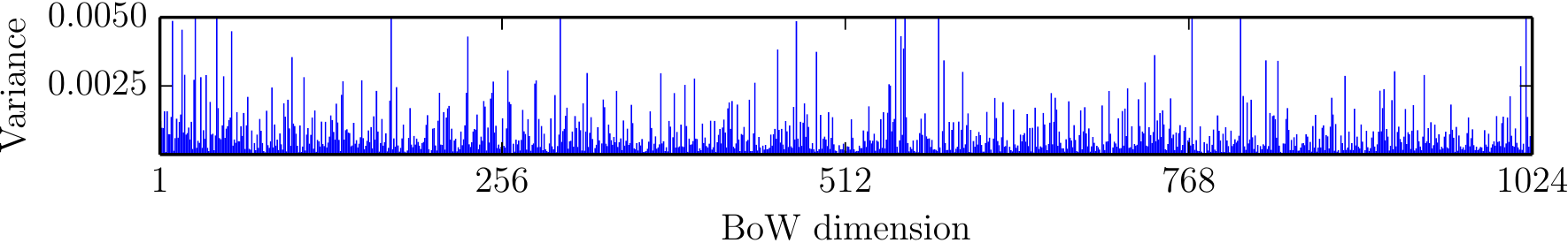}
    \caption{$\ell_2$ normalized D-BoTSW}
\end{subfigure}
\begin{subfigure}[b]{0.8\textwidth}
	\centering
	\includegraphics[width=\linewidth]{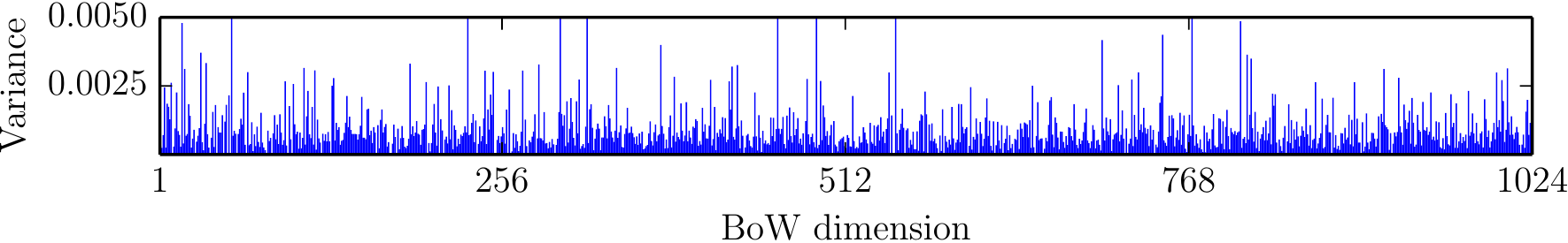}
    \caption{IDF+$\ell_2$ normalized D-BoTSW}
\end{subfigure}
\begin{subfigure}[b]{0.8\textwidth}
	\centering
	\includegraphics[width=\linewidth]{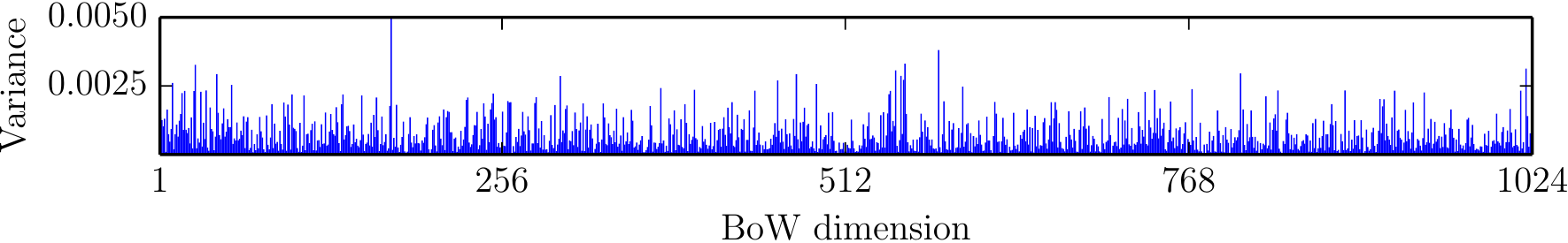}
    \caption{SSR+$\ell_2$ normalized D-BoTSW}
\end{subfigure}
\caption{Per-dimension energy of D-BoTSW vectors extracted from dataset \emph{ShapesAll}. The same codebook is used for all normalization schemes so that dimensions are comparable across all three sub-figures. \label{fig:var_dense_norm}}
\end{figure}

This is backed by~\figurename~\ref{fig:var_dense_norm}, in which one can see that when using SSR normalization, variance (\emph{i.e.} energy) is spread across all dimensions of the BoW, leading to a more balanced representation than with other two normalization schemes. 

\subsection{Comparison with state-of-the-art methods}

In the following, we will refer to dense SSR-normalized BoTSW as D-BoTSW, since this setup is the one providing the best classification performance. 
We now compare D-BoTSW to the most popular state-of-the-art methods for time series classification. 
The UCR repository provides error rates for the 86 datasets with Euclidean distance 1NN (EDNN) and Dynamic Time Warping 1NN (DTWNN)~\cite{ratanamahatana2004everything}. 
We use published error rates for TSBF (45 datasets)~\cite{baydogan2013bof}, SAX-VSM (51 datasets)~\cite{senin2013saxsvm}, SMTS (45 datasets)~\cite{Bay15}, PROP (46 datasets)~\cite{lines14} and BoP (20 datasets).

As BoP~\cite{lin2012bop} only provides classification performance for 20 datasets, we decided not to plot pairwise comparison of error rates between D-BoTSW and BoP.
Note however that the Win/Tie/Lose score is 17/1/2 in favor of D-BoTSW and this difference is statistically significant ($p<0.001$). BoP has smaller error rate than D-BoTSW on \emph{wafer} (0.003 \emph{vs.} 0.004) and \emph{Olive Oil} (0.133 \emph{vs.} 0.167) data sets. 



\begin{figure}[!p]
\centering
	\includegraphics[width=0.48\textwidth]{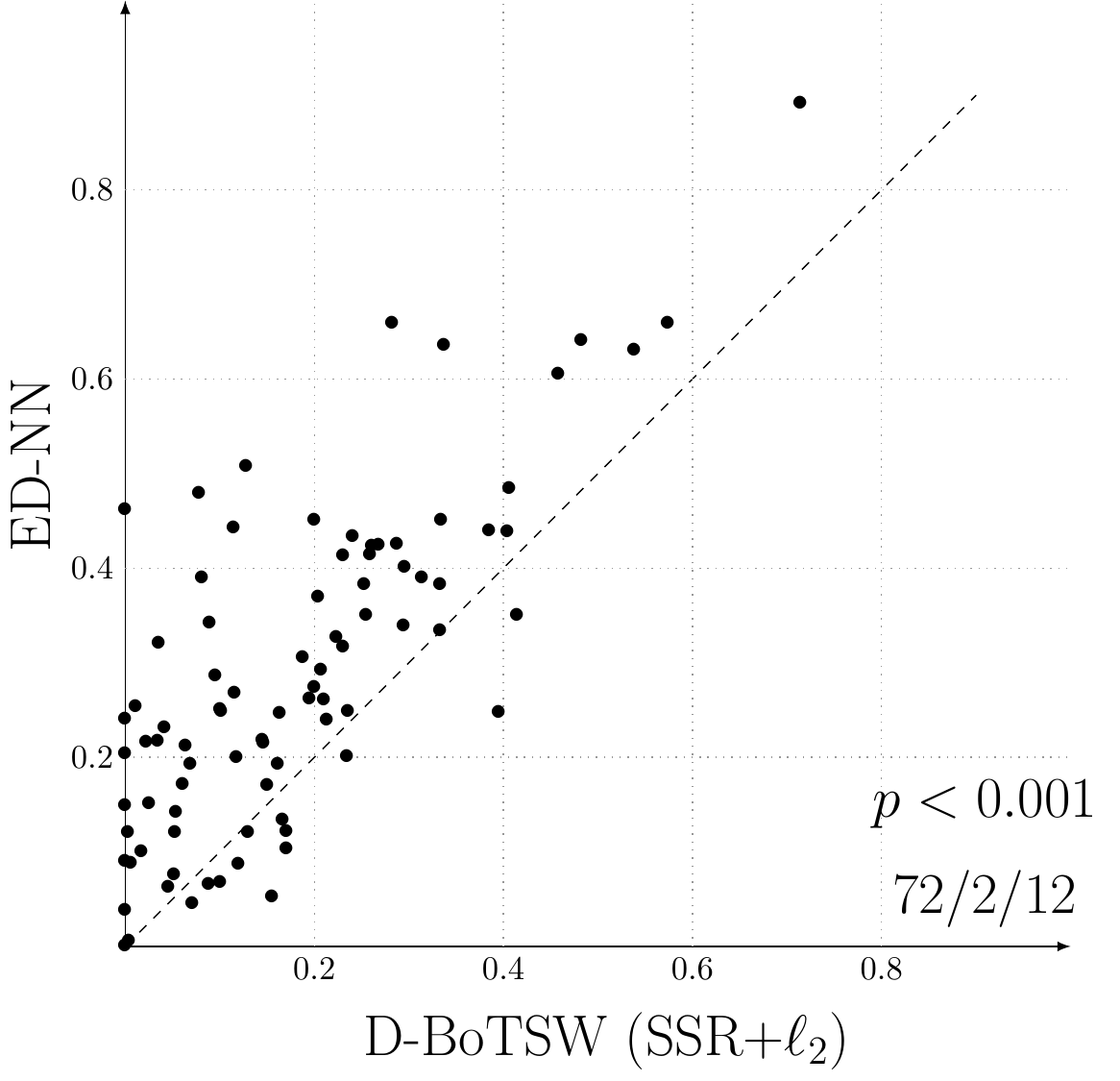}
	\includegraphics[width=0.48\textwidth]{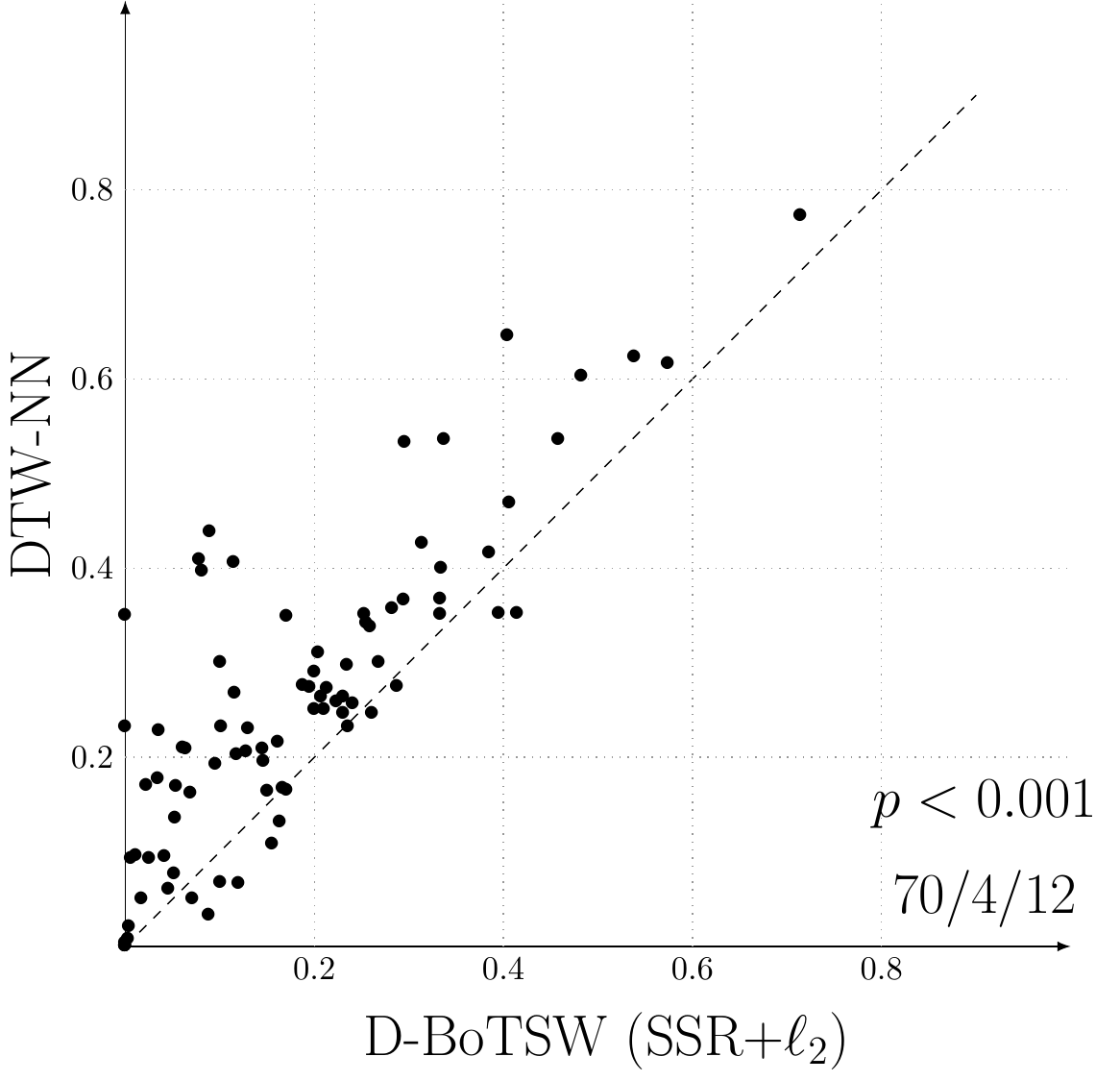}
	\includegraphics[width=0.48\textwidth]{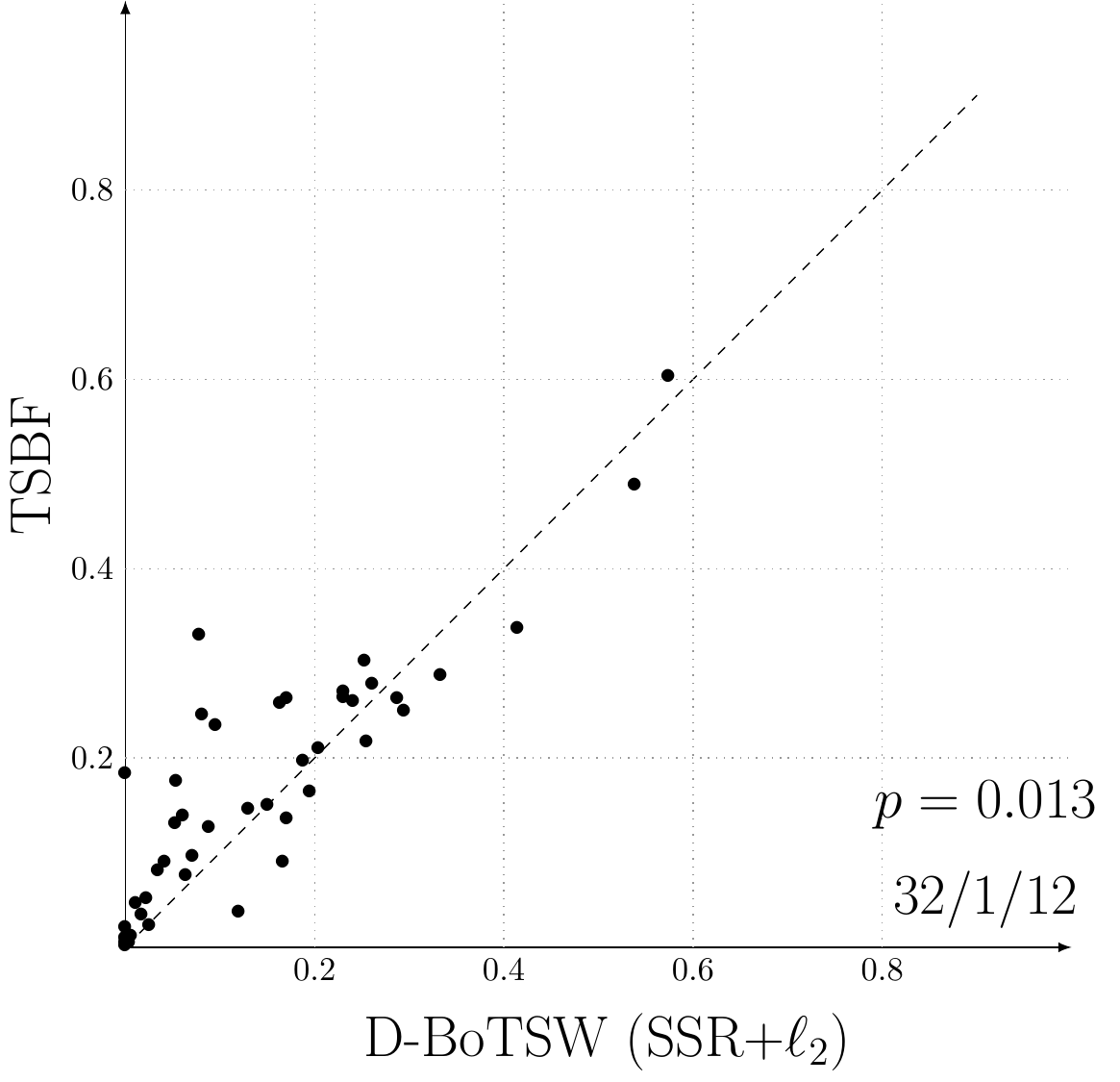}
	\includegraphics[width=0.48\textwidth]{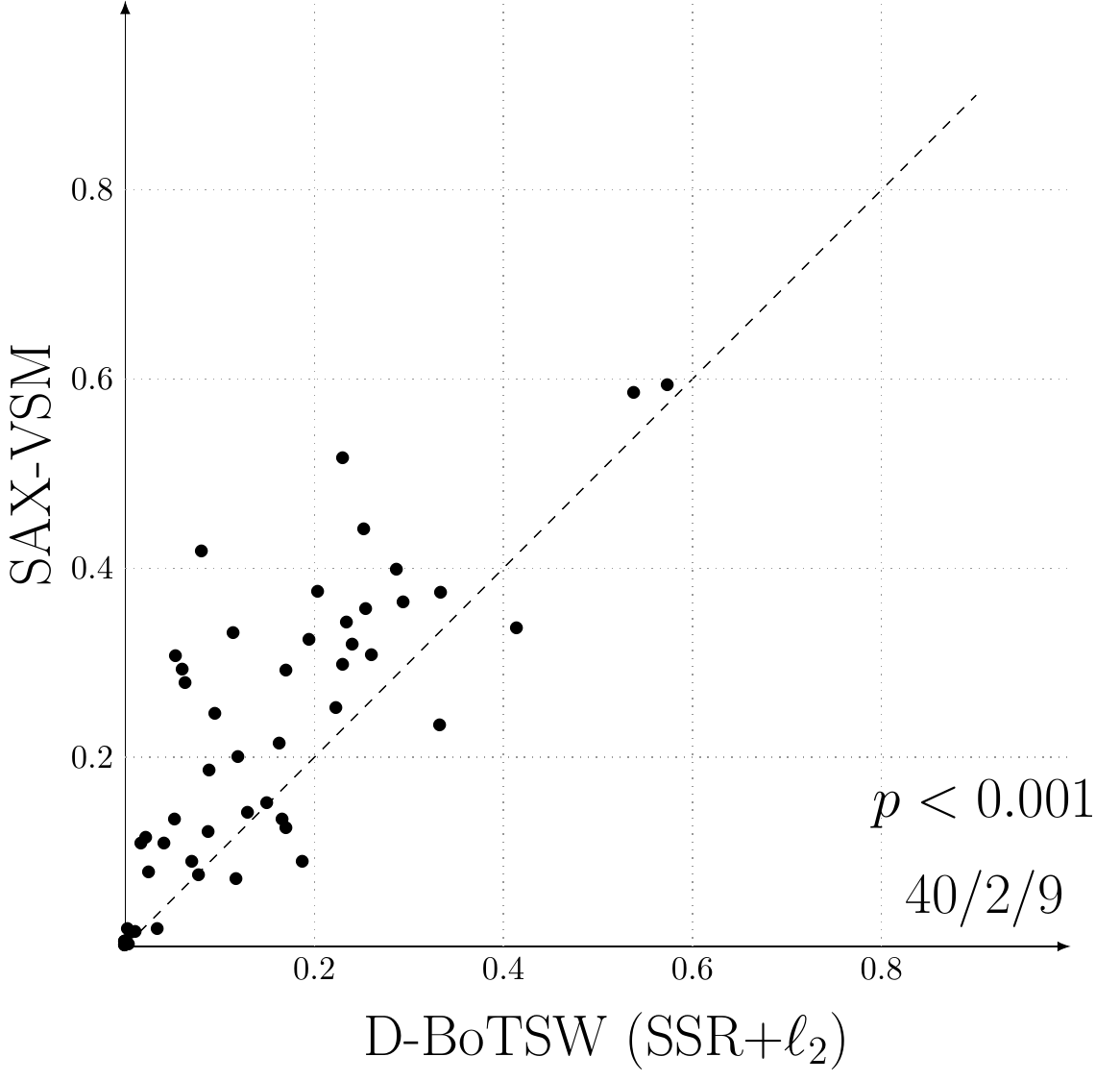}
	\includegraphics[width=0.48\textwidth]{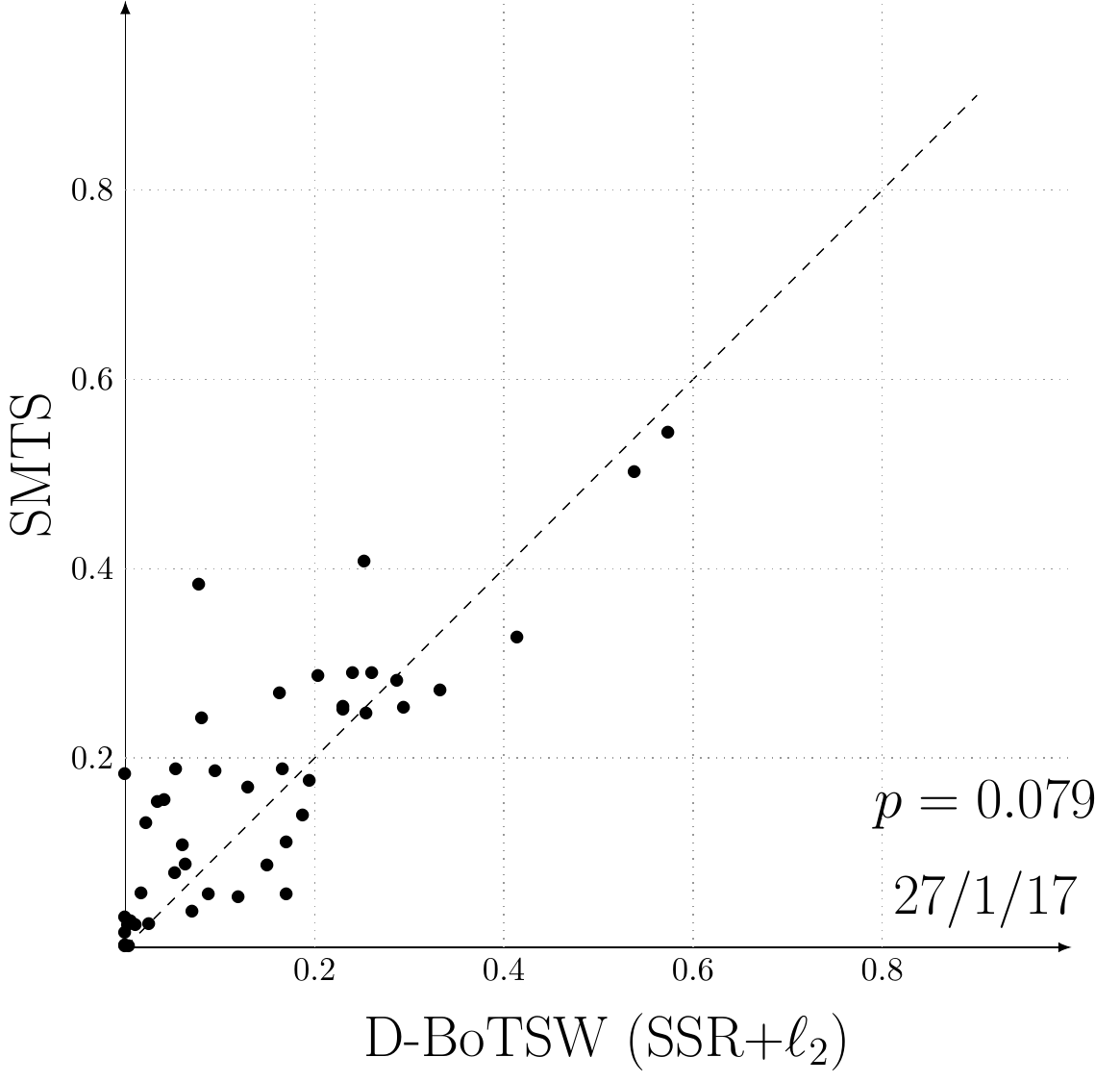}
	\includegraphics[width=0.48\textwidth]{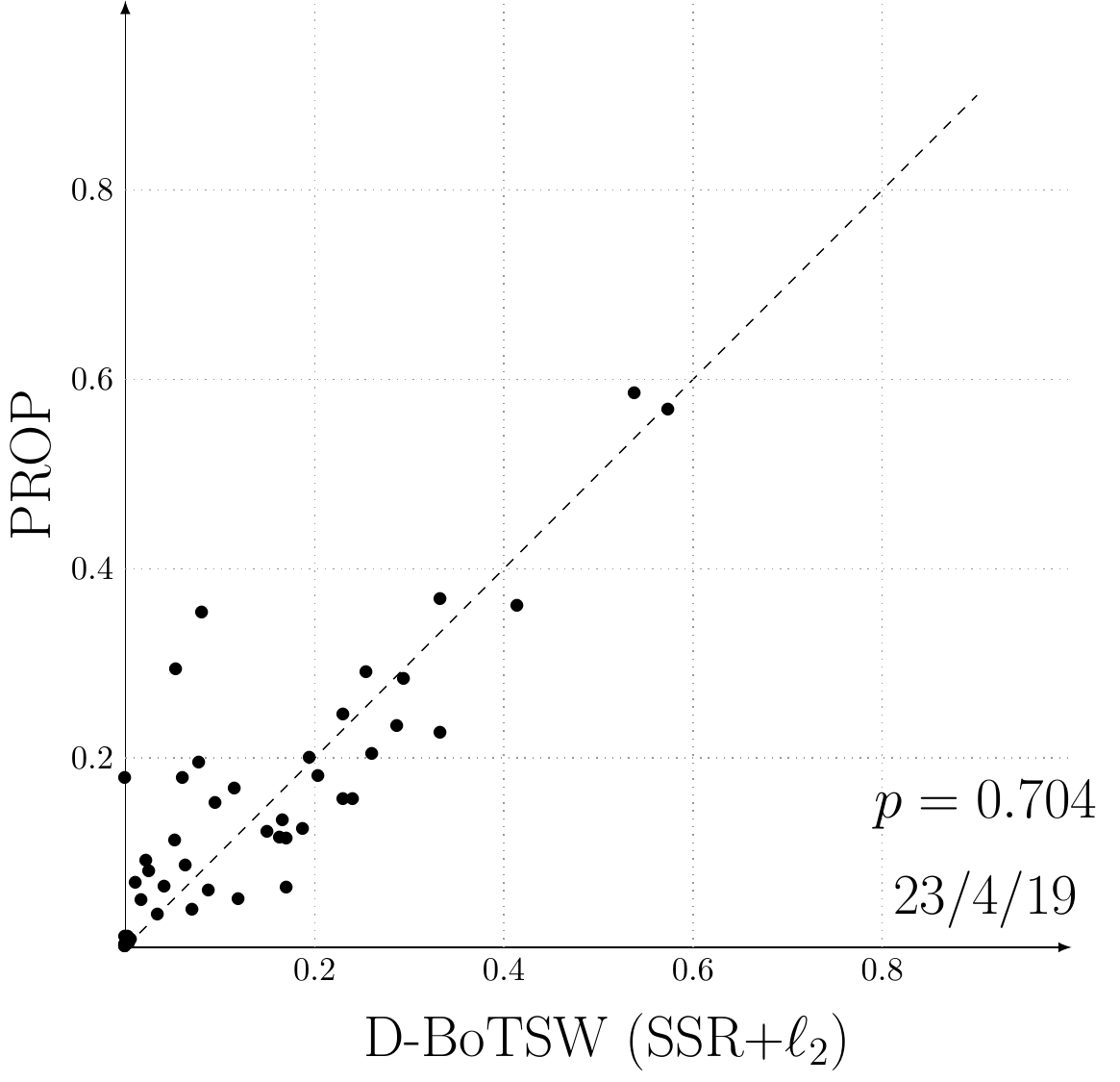}
\caption{\label{fig:errorrate_baselines}
Error rates for D-BoTSW with SSR normalization versus baselines (ED-NN, DTW-NN, TSBF, SAX-VSM, SMTS, PROP).  
}
\end{figure}


\begin{table}[!p]
	\centering
		\resizebox{0.47\textwidth}{!}{
\begin{tabular}{c|ccc}
Dataset & EDNN & DTWNN & D-BoTSW\\
\hline
50words & 0.369 & 0.31 & \bf 0.204\\
Adiac & 0.389 & 0.396 & \bf 0.082\\
ArrowHead & \bf 0.2 & 0.297 &  0.234\\
Beef & \bf 0.333 & 0.367 & \bf 0.333\\
BeetleFly & 0.25 & 0.3 & \bf 0.1\\
BirdChicken & 0.45 & 0.25 & \bf 0.2\\
Car & 0.267 & 0.267 & \bf 0.117\\
CBF & 0.148 & 0.003 & \bf 0\\
\begin{tabular}{c}Chlorine\\Concentration\end{tabular} & \bf 0.35 & 0.352 & 0.414\\
\begin{tabular}{c}CinC\_ECG\\\_torso\end{tabular} & \bf 0.103 & 0.349 & 0.17\\
Coffee & \bf 0 & \bf 0 & \bf 0\\
Computers & 0.424 & 0.3 & \bf 0.268\\
Cricket\_X & 0.423 & \bf 0.246 & 0.262\\
Cricket\_Y & 0.433 & 0.256 & \bf 0.241\\
Cricket\_Z & 0.413 & 0.246 & \bf 0.231\\
\begin{tabular}{c}DiatomSize\\Reduction\end{tabular} & 0.065 & \bf 0.033 & 0.088\\
\begin{tabular}{c}DistalPhalanx\\OutlineAgeGroup\end{tabular} & 0.218 & 0.208 & \bf 0.145\\
\begin{tabular}{c}DistalPhalanx\\OutlineCorrect\end{tabular} & 0.248 & \bf 0.232 & 0.235\\
DistalPhalanxTW & 0.273 & 0.29 & \bf 0.2\\
Earthquakes & 0.326 & 0.258 & \bf 0.224\\
ECG200 & \bf 0.12 & 0.23 & 0.13\\
ECG5000 & 0.075 & 0.076 & \bf 0.052\\
ECGFiveDays & 0.203 & 0.232 & \bf 0\\
ElectricDevices & 0.45 & 0.399 & \bf 0.334\\
FaceAll & 0.286 & 0.192 & \bf 0.095\\
FaceFour & 0.216 & 0.17 & \bf 0.023\\
FacesUCR & 0.231 & 0.095 & \bf 0.041\\
FISH & 0.217 & 0.177 & \bf 0.034\\
FordA & 0.341 & 0.438 & \bf 0.089\\
FordB & 0.442 & 0.406 & \bf 0.116\\
Gun\_Point & 0.087 & 0.093 & \bf 0.007\\
Ham & 0.4 & 0.533 & \bf 0.295\\
HandOutlines & 0.199 & 0.202 & \bf 0.119\\
Haptics & 0.630 & 0.623 & \bf 0.539\\
Herring & 0.484 & 0.469 & \bf 0.406\\
InlineSkate & 0.658 & 0.616 & \bf 0.575\\
\begin{tabular}{c}Insect\\WingbeatSound\end{tabular} & 0.438 & 0.645 & \bf 0.405\\
ItalyPowerDemand & \bf 0.045 & 0.05 & 0.072\\
\begin{tabular}{c}LargeKitchen\\Appliances\end{tabular} & 0.507 & 0.205 & \bf 0.128\\
Lightning2 & 0.246 & \bf 0.131 & 0.164\\
Lightning7 & 0.425 & \bf 0.274 & 0.288\\
MALLAT & 0.086 & \bf 0.066 & 0.12\\
Meat & \bf 0.067 & \bf 0.067 & 0.1\\
MedicalImages & 0.316 & 0.263 & \bf 0.23\\
\begin{tabular}{c}MiddlePhalanx\\OutlineAgeGroup\end{tabular} & 0.26 & 0.25 & \bf 0.21\\
\end{tabular}
}
\resizebox{0.47\textwidth}{!}{
\begin{tabular}{c|ccc}
Dataset & EDNN & DTWNN & D-BoTSW\\
\hline
\begin{tabular}{c}MiddlePhalanx\\OutlineCorrect\end{tabular} & \bf 0.247 & 0.352 & 0.395\\
MiddlePhalanxTW & 0.439 & 0.416 & \bf 0.386\\
MoteStrain & \bf 0.121 & 0.165 & 0.17\\
\begin{tabular}{c}NonInvasiveFetal\\ECG\_Thorax1\end{tabular} & 0.171 & 0.209 & \bf 0.061\\
\begin{tabular}{c}NonInvasiveFetal\\ECG\_Thorax2\end{tabular} & 0.12 & 0.135 & \bf 0.053\\
OliveOil & \bf 0.133 & 0.167 & 0.167\\
OSULeaf & 0.479 & 0.409 & \bf 0.079\\
\begin{tabular}{c}PhalangesOutlines\\Correct\end{tabular} & 0.239 & 0.272 & \bf 0.213\\
Phoneme & 0.891 & 0.772 & \bf 0.714\\
Plane & 0.038 & \bf 0 & \bf 0\\
\begin{tabular}{c}ProxiamlPhalanx\\OutlineAgeGroup\end{tabular} & 0.215 & 0.195 & \bf 0.146\\
\begin{tabular}{c}ProxiamlPhalanx\\OutlineCorrect\end{tabular} & 0.192 & 0.216 & \bf 0.162\\
ProximalPhalanxTW & 0.292 & 0.263 & \bf 0.208\\
RefrigerationDevices & 0.605 & 0.536 & \bf 0.459\\
ScreenType & 0.64 & 0.603 & \bf 0.483\\
ShapeletSim & 0.461 & 0.35 & \bf 0\\
ShapesAll & 0.248 & 0.232 & \bf 0.102\\
\begin{tabular}{c}SmallKitchen\\Appliances\end{tabular} & 0.659 & 0.357 & \bf 0.283\\
\begin{tabular}{c}SonyAIBORobot\\Surface\end{tabular} & 0.141 & 0.169 & \bf 0.055\\
\begin{tabular}{c}SonyAIBORobot\\SurfaceII\end{tabular} & 0.305 & 0.275 & \bf 0.189\\
StarLightCurves & 0.151 & 0.093 & \bf 0.026\\
Strawberry & 0.062 & 0.06 & \bf 0.046\\
SwedishLeaf & 0.211 & 0.208 & \bf 0.064\\
Symbols & 0.1 & 0.05 & \bf 0.017\\
synthetic\_control & 0.12 & 0.007 & \bf 0.003\\
ToeSegmentation1 & 0.320 & 0.228 & \bf 0.035\\
ToeSegmentation2 & 0.192 & 0.162 & \bf 0.069\\
Trace & 0.24 & \bf 0 & \bf 0\\
Two\_Patterns & 0.09 & \bf 0 & \bf 0\\
TwoLeadECG & 0.253 & 0.096 & \bf 0.011\\
\begin{tabular}{c}uWaveGesture\\Library\_X\end{tabular} & 0.261 & 0.273 & \bf 0.195\\
\begin{tabular}{c}uWaveGesture\\Library\_Y\end{tabular} & 0.338 & 0.366 & \bf 0.294\\
\begin{tabular}{c}uWaveGesture\\Library\_Z\end{tabular} & 0.35 & 0.342 & \bf 0.255\\
\begin{tabular}{c}uWaveGesture\\LibraryAll\end{tabular} & \bf 0.052 & 0.108 & 0.156\\
wafer & 0.005 & 0.02 & \bf 0.004\\
Wine & 0.389 & 0.426 & \bf 0.315\\
WordsSynonyms & 0.382 & 0.351 & \bf 0.254\\
WordSynonyms & 0.382 & 0.351 & \bf 0.334\\
Worms & 0.635 & 0.536 & \bf 0.337\\
WormsTwoClass & 0.414 & 0.337 & \bf 0.26\\
yoga & 0.170 & 0.164 & \bf 0.15\\
~ & \\
\end{tabular}
}

	\caption{\label{tab:errorrate}
	Classification error rates for D-BoTSW with SSR normalization (for each dataset, best performance is written as bold text).}
\end{table}

\figurename~\ref{fig:errorrate_baselines} shows that D-BoTSW performs better than 1NN combined with ED (EDNN) or DTW (DTWNN), TSBF, SAX-VSM and SMTS. 
Though relying on a single similarity measure that has linear time complexity in the length of time series, D-BoTSW slightly outperforms PROP, which relies on outputs from several similarity measures with quadratic time complexity.
In~\figurename~\ref{fig:errorrate_baselines}, it is striking to realize that D-BoTSW not only improves the classification, but might improve it considerably. Error rate on \emph{Shapelet Sim} dataset drops from 0.461 (EDNN) and 0.35 (DTWNN) to 0 (D-BoTSW), for example.
Pairwise comparisons of methods 
show that all observed differences between D-BoTSW and state-of-the-art methods are statistically significant, except for PROP.
Error rates (ER) obtained with D-BoTSW are reported in~\tablename~\ref{tab:errorrate}, together with baseline scores publicly available at~\cite{ucr}. 

This set of experiments, conducted on a wide variety of time series datasets, shows that D-BoTSW significantly outperforms most state-of-the-art methods.

\section{Conclusion}
\label{sec:ccl}

In this paper, we presented the D-BoTSW technique, which transforms time series into histograms of quantized local features. 
The association of SIFT keypoints and Bag-of-Words has been widely used and is considered as a standard technique in image domain, however it has never been investigated for time series classification.
We carried out extensive experiments and showed that dense keypoint extraction and SSR normalization of Bag-of-Words lead to the best performance for our method.
We compared the results with standard techniques for time series classification: D-BoTSW has comparable performance to PROP with lower time complexity and significantly outperforms all other techniques.

We believe that classification performance could be further improved by taking more time information into account, as well as reducing the impact of quantization losses in our representation. 
Indeed, only local temporal information is embedded in our model and the global structure of time series is ignored.
Moreover, more detailed global representations for sets of features than the standard BoW have been proposed in the computer vision community~\cite{jegou2010aggregating,perronnin2007fisher}, and such global features could be used in our framework.

\section*{Acknowledgments}

This work has been partly funded by ANR project ASTERIX (ANR-13-JS02-0005-01), R{\'e}gion Bretagne and CNES-TOSCA project VEGIDAR.

\bibliography{main}{}
\bibliographystyle{plain}
\end{document}